\documentclass[]{bytedance_seed}

\usepackage[toc,page,header]{appendix}

\usepackage{minitoc}
\usepackage{amsmath}
\usepackage{amssymb}
\usepackage{hyperref}
\usepackage[most]{tcolorbox}
\usepackage{listings}
\usepackage{xcolor}
\usepackage{listings}
\definecolor{jsonkey}{rgb}{0.36,0.54,0.66}
\definecolor{jsonstring}{rgb}{0.58,0.36,0.18}
\definecolor{jsonnumber}{rgb}{0.2,0.2,0.2}

\lstdefinelanguage{json}{
  basicstyle=\ttfamily\footnotesize,
  showstringspaces=false,
  breaklines=true,
  frame=none,
  string=[b]",
  stringstyle=\color{jsonstring},
  keywordstyle=\color{jsonkey},
  morekeywords={true,false,null},
  literate=
   *{0}{{\color{jsonnumber}0}}1
    {1}{{\color{jsonnumber}1}}1
    {2}{{\color{jsonnumber}2}}1
    {3}{{\color{jsonnumber}3}}1
    {4}{{\color{jsonnumber}4}}1
    {5}{{\color{jsonnumber}5}}1
    {6}{{\color{jsonnumber}6}}1
    {7}{{\color{jsonnumber}7}}1
    {8}{{\color{jsonnumber}8}}1
    {9}{{\color{jsonnumber}9}}1
}

\newcommand{\ours}{\textit{StoryMem}}
\newcommand{\ie}{\textit{i.e.}}
\newcommand{\eg}{\textit{e.g.}}

\title{StoryMem: Multi-shot Long Video \\ Storytelling with Memory}

\author[1,2,*]{Kaiwen Zhang}
\author[2, \dagger]{Liming Jiang}
\author[2]{Angtian Wang}
\author[2]{Jacob Zhiyuan Fang}
\author[2]{Tiancheng Zhi}
\author[2]{Qing Yan}
\author[2]{Hao Kang}
\author[2]{Xin Lu}
\author[1,\S]{Xingang Pan}

\affiliation[1]{S-Lab, Nanyang Technological University}
\affiliation[2]{Intelligent Creation, ByteDance}

\contribution[*]{Work done during internship at ByteDance}
\contribution[\dagger]{Project Lead}
\contribution[\S]{Corresponding Author}

\abstract{
Visual storytelling requires generating multi-shot videos with cinematic quality and long-range consistency.
Inspired by human memory, we propose~\textbf{\ours}, a paradigm that reformulates long-form video storytelling as iterative shot synthesis conditioned on explicit visual memory, transforming pre-trained single-shot video diffusion models into multi-shot storytellers.
This is achieved by a novel \textbf{Memory-to-Video (M2V)} design, which maintains a compact and dynamically updated memory bank of keyframes from historical generated shots. 
The stored memory is then injected into single-shot video diffusion models via latent concatenation and negative RoPE shifts with only LoRA fine-tuning.
A semantic keyframe selection strategy, together with aesthetic preference filtering,  further ensures informative and stable memory throughout generation.
Moreover, the proposed framework naturally accommodates smooth shot transitions and customized story generation application.
To facilitate evaluation, we introduce ST-Bench, a diverse benchmark for multi-shot video storytelling.
Extensive experiments demonstrate that~\ours~achieves superior cross-shot consistency over previous methods while preserving high aesthetic quality and prompt adherence, marking a significant step toward coherent minute-long video storytelling.
}

\date{\today}

\checkdata[Project Page]{\url{https://kevin-thu.github.io/StoryMem}}

\begin{document}
\maketitle


\section{Introduction}
\label{sec:intro}

\begin{figure}[t]
    \vspace{-0.5cm}
    \centering
    \includegraphics[width=0.98\linewidth]{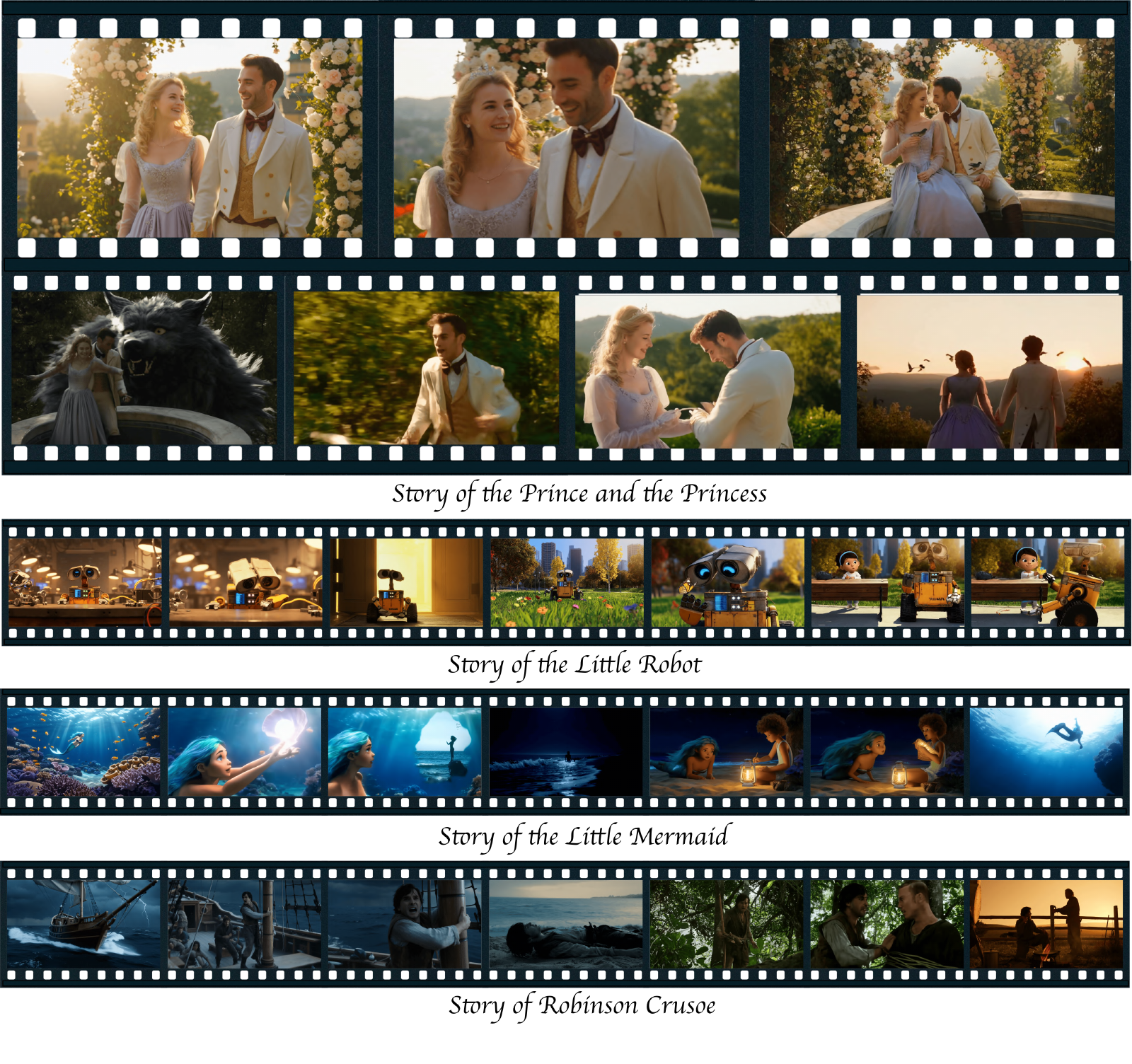}
    \vspace{-0.75cm}
    \caption{Given a story script with per-shot text descriptions, \ours~generates appealing minute-long, multi-shot narrative videos with highly coherent characters and cinematic visual quality. This is achieved through shot-by-shot generation using a memory-conditioned single-shot video diffusion model.}
    \vspace{-0.3cm}
\label{fig:teaser}
\end{figure}

Storytelling is a core expression of human creativity, spanning from cave art to cinema. 
Recent years have witnessed substantial advances in video diffusion models~\cite{ddpm, ldm, Blattmann2023SVD}, enabling the synthesis of single-shot short videos with near-cinematic visual fidelity~\cite{sora, sora2, kling, veo, wan2025}.
A natural next step is to move beyond isolated clips and enable video models to construct coherent visual narratives.
However, genuine storytelling demands multi-layered coherence across shots and scenes, from low-level consistency of characters and environments to high-level alignment of visual style and narrative flow. 
Achieving such minute-long, multi-shot narrative generation remains a significant challenge.

Existing solutions fall into two main categories. 
The \textit{first} category models all shots jointly within a single large video diffusion model.  
LCT~\cite{Guo2025LCT}, for example, employs full attention with interleaved 3D RoPE~\cite{Su2021RoPE} to capture cross-shot dependencies, but incurs quadratic training and inference costs as sequence length grows.  
Follow-up methods improve efficiency via token compression~\cite{Xiao2025CaptainCT} or sparse attention~\cite{Cai2025MOC, jia2025moga, meng2025holocine}, yet still require large-scale retraining on multi-shot videos and often degrade visual quality relative to their pretrained single-shot base models.
The \textit{second} one adopts a keyframe-based decoupled pipeline~\cite{Hu2024StoryAgentCS, Zhao2024MovieDreamerHG, Wu2025MovieAgent, Zhuang2024VloggerMY, Long2024VideoStudioGC}: the first frame of each shot is generated using in-context image models~\cite{Zhou2024StoryDiffusion, Tewel2024Consistory, Huang2024ICLora} and then expanded into a clip with pretrained image-to-video (I2V) models. 
Although efficient and leveraging high-quality single-shot models, this independent-shot design lacks temporal awareness, \ie, no context propagates across shots, leading to inconsistent visual details, disconnected scene evolution, and rigid transitions.
Consequently, changes such as new character appearances or shifting camera viewpoints cannot be coherently maintained throughout the long video.

These two lines of research reflect a core dilemma in narrative video generation: joint training demand heavy computation and scarce high-quality multi-shot data, while decoupled approaches suffer from inconsistency. 
In this work, we explore a third path that achieves both high consistency and high efficiency.
Our key insight is that \textit{high-quality pretrained single-shot video diffusion models can be effectively adapted for long-term coherent storytelling when augmented with visual memory.}
Inspired by how human memory selectively retains salient visual information, we introduce \textbf{\ours}, a new paradigm that reformulates long-form video storytelling as iterative shot synthesis conditioned on past keyframes and per-shot descriptions. 
Central to this paradigm is our \textbf{Memory-to-Video (M2V)} design, a lightweight framework that injects explicit visual memory into single-shot video diffusion models to enforce cross-shot consistency.
%
Instead of training a single monolithic model on long-video data, \ours~generates stories shot by shot via maintaining a compact memory bank that stores key visual context, \eg, characters, scenes, and stylistic cues, from previously generated shot.
The memory then serves as a global conditioning signal, guiding subsequent generation toward consistent visual semantics and smooth scene evolution (see~Fig.~\ref{fig:teaser}).

To realize memory-conditioned generation, we introduce a simple yet effective design that combines memory latent concatenation with a negative RoPE shift, naturally extending pretrained I2V models to capture long-term dependencies and maintain context-level consistency. 
We further propose a semantic keyframe selection strategy based on CLIP~\cite{clip} features, along with aesthetic preference filtering via HPSv3~\cite{hpsv3}, to preserve a compact memory bank that is both informative and reliable. 
During generation, the memory is dynamically extracted, updated, and injected into the model to guide each new shot. 
Notably, \ours~requires only LoRA~\cite{Hu2021LoRA} fine-tuning on semantically coherent short video clips, achieving strong cross-shot consistency without compromising the high visual quality of pretrained single-shot video diffusion models. 
Beyond text-driven story generation, \ours~serves as a versatile framework applicable to broader paradigms, enabling natural scene transitions when combined with I2V control, and supporting customized reference-to-video (R2V) generation by incorporating reference images into the memory.


To support evaluation and comparison, we introduce \textit{ST-Bench}, a multi-scene, multi-shot video storytelling benchmark covering diverse complex narratives.
Extensive experiments show that \ours~achieves superior cross-shot consistency while retaining high visual fidelity and strong prompt adherence, surpassing state-of-the-art methods. 
Our approach marks a significant step toward coherent minute-long video storytelling, bridging the gap between single-shot video diffusion and long-form visual storytelling.

\section{Related Work}
\label{sec:relwork}

\noindent\textbf{Keyframe-based Story Generation.}
Prior research on visual storytelling primarily focuses on 
story image generation~\cite{Li2018StoryGAN, Rahman2022MakeAStory, Zhou2024StoryDiffusion, Liu2023StoryGen, Tewel2024Consistory, Mao2024StoryAdapterAT, Yang2024SEEDStoryML, He2024DreamStoryOS, Dinkevich2025Story2BoardAT, Wang2025CharaConsistFC}.
For instance, StoryDiffusion~\cite{Zhou2024StoryDiffusion} introduces Consistent Self-Attention into a text-to-image (T2I) model~\cite{ddpm, ldm} to produce character-consistent storyboards.
With the emergence of video generation models~\cite{sora, kling, veo, sora2, Blattmann2023SVD, Yang2024CogVideoXTD, Kong2024HunyuanVideoAS, wan2025}, subsequent works~\cite{Hu2024StoryAgentCS, Zhao2024MovieDreamerHG, Wu2025MovieAgent, Zhuang2024VloggerMY, Long2024VideoStudioGC} extend image pipelines to videos through an agentic, keyframe-based framework. 
These methods~\cite{Hu2024StoryAgentCS, Zhao2024MovieDreamerHG, Wu2025MovieAgent, Zhuang2024VloggerMY, Long2024VideoStudioGC} first generate keyframes using story image models or in-context image editing models~\cite{Huang2024ICLora, Labs2025FLUXKontext, nanobanana}, then expand them into video clips with image-to-video (I2V) models. However, consistency is guaranteed only at the keyframe level, leaving each shot largely independent and transitions between shots often rigid.
In contrast, we introduce a memory-to-video (M2V) framework that generates coherent story videos in a shot-by-shot manner, while preserving contextual information across shots with minimal computational overhead.

\noindent\textbf{Multi-shot Long Video Generation.}
Beyond keyframe-based storytelling, a parallel line of research targets general multi-shot long video generation, by fine-tuning pretrained single-shot models (\eg, Wan~\cite{wan2025}).
LCT~\cite{Guo2025LCT} pioneers this direction by jointly modeling multi-shot videos using full attention~\cite{Vaswani2017AttentionIA} and interleaved 3D RoRE~\cite{Su2021RoPE}, achieving strong cross-shot coherence but incurring quadratic computational cost.
Recently, a line of concurrent works follow LCT's framework and improve efficiency via token compression~\cite{Xiao2025CaptainCT} or sparse attention~\cite{Cai2025MOC, jia2025moga, meng2025holocine}.
However, they require large-scale training on multi-shot long video data, often resulting in quality degradation compared to their pretrained high-quality single-shot base model.
In contrast, our M2V framework requires only lightweight LoRA~\cite{Hu2021LoRA} fine-tuning on short video clips, thereby preserving the visual fidelity of the base model while seamlessly adaptable to other paradigms such as image-to-video (I2V) and reference-to-video (R2V) for customized story generation.

\noindent\textbf{Memory Mechanisms in Video Generation.}
Memory is a fundamental capability of intelligent systems, and has been widely studied in large language models~\cite{Zhang2024memorysurvey, Du2025memorysurvey, Wu2025memorysurvey}. In contrast, memory mechanisms for video generation remain largely unexplored.
Existing attempts introduce memory into video world models through inference-time training~\cite{Hong2024SlowFastVGenSL}, 3D modeling~\cite{Li2025VMemCI, huang2025memoryforcing}, or camera-based memory frame retrieval and cross attention~\cite{Xiao2025WORLDMEMLC, Yu2025ContextAM}.
However, they primarily target spatial consistency under controllable world simulations and rely on auxiliary control inputs such as actions or camera poses, limiting the applicability to general-purpose video generation.
To the best of our knowledge, we are the first to introduce an explicit memory mechanism into general video generation models. 
This enables the model to memorize essential characters and background scenes throughout the evolving generation process, thus maintaining consistency across minute-long generation.
\section{Methodology}
\label{sec:method}

\begin{figure*}[t]
    \centering
    \includegraphics[width=\linewidth]{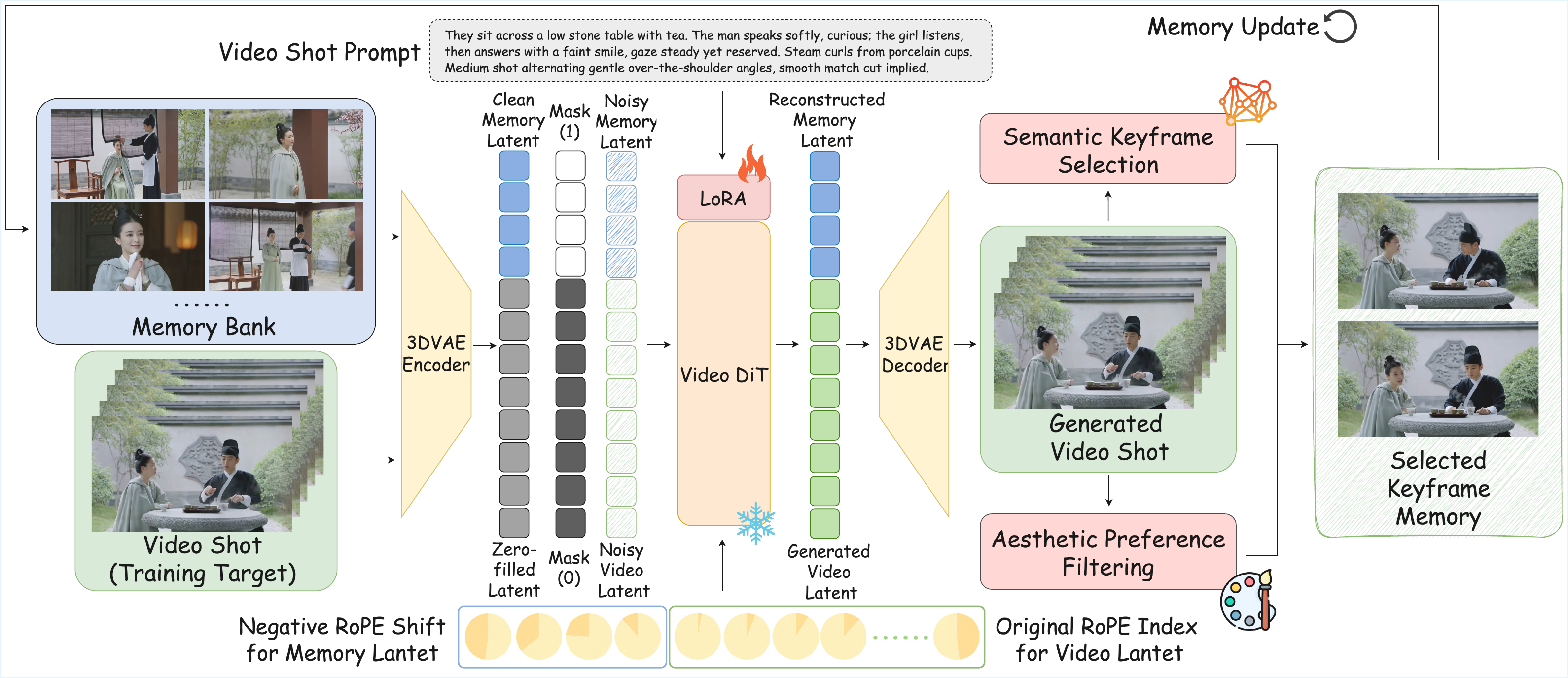}
    \caption{\textbf{Overview of~\ours.} \ours~generates each shot conditioned on a memory bank that stores keyframes from previously generated shots. During generation, the selected memory frames are encoded by a 3D VAE, fused with noisy video latents and binary masks, and fed into a LoRA-finetuned memory-conditioned Video DiT to synthesize the current shot. After generating each shot, semantic keyframe selection and aesthetic preference filtering are applied to obtain informative and reliable memory frames, enabling long-range cross-shot consistency and natural narrative progression. By iteratively generating shots with memory updates, \ours~produces coherent minute-long, multi-shot story videos.}
    \vspace{-0.3cm}
\label{fig:method}
\end{figure*}

In this section, we present \ours, a novel pipeline that builds upon single-shot video diffusion models to address the challenge of coherent multi-shot story generation.
We first introduce preliminaries in Sec.~\ref{subsec:prelim}.
Then, we formalize the problem in Sec.~\ref{subsec:formulation}, and introduce our proposed Memory-to-Video (M2V) mechanism in Sec.~\ref{subsec:m2v}, which enforces cross-shot consistency via memory-conditioned video generation.
Sec.~\ref{subsec:mfextract} further presents our memory extraction and update strategy for applying the finetuned M2V model to long multi-shot storytelling.
Finally, Sec.~\ref{subsec:m2vext} extends M2V to MI2V and MR2V for natural scene transitions and customized control.
An overview of~\ours~is shown in Fig.~\ref{fig:method}.

\subsection{Preliminary}
\label{subsec:prelim}
\noindent\textbf{Video Diffusion Model.}
Our method is built upon a latent video diffusion model~\cite{ddpm, ldm, wan2025}.
The diffusion process operates on the video latents $z_0 = \mathcal{E}(v)\in \mathbb{R}^{c\times f\times h\times w}$ by encoding RGB video $v$ with 3D VAE~\cite{vae} encoder $\mathcal{E}$.
The diffusion model learns the distribution of video latents by learning to transform the noise samples $z_1=\epsilon \sim \mathcal{N}(\mathbf{0}, \mathbf{I})$ to target data sample $z_0$ in terms of a differential equation:
\begin{equation}
dz_t = v_{\Theta}(z_t, t)\,dt, \quad t \in [0, 1],
\end{equation}
where velocity field $v_\Theta$ is parametrized by a neural network $\Theta$.
The network is trained with a velocity prediction loss under the rectified flow~\cite{Liu2022FlowSA, Lipman2022FlowMF} formulation:
\begin{equation}
z_t = (1 - t)z_0 + t\epsilon,\quad \epsilon \sim \mathcal{N}(\mathbf{0}, \mathbf{I}),
\end{equation}
\vspace{-5mm}
\begin{equation}
\label{eq:rf}
\mathcal{L}_{RF} = \mathbb{E}_{z_0,\, \epsilon,\, t} \left[ \left\| v_{\Theta}(z_t, t) - (z_0 - \epsilon) \right\|^2 \right].
\end{equation}

\noindent\textbf{Model Architecture.}
To reach cinematic-level quality, we adopt state-of-the-art single-shot video diffusion model \textit{Wan2.2-I2V}~\cite{wan2025} as our base model, which uses diffusion transformer (DiT)~\cite{dit} as the velocity prediction network $\Theta$.
Each DiT block contains self-attention for intra-video dependency modeling and cross-attention for text conditioning, and uses 3D Rotary Position Embedding (RoPE)~\cite{Su2021RoPE} to encode spatial and temporal coordinates.
Wan-I2V model further conditions on an first-frame image to guide video synthesis.
The image is concatenated with zero-filled frames and encoded by the 3D VAE into a conditional latent $z_c\in\mathbb{R}^{c\times f\times h\times w}$.
A binary mask $M\in\{0,1\}^{s\times f\times h\times w}$ indicates which frames are preserved or generated, where $s$ denotes the temporal stride of the 3D VAE, and $F=s\times f$ is the original frame count of the raw video.
During the diffusion process, the noisy latent $z_t$, the conditional latent $z_c$, and the mask $M$ are concatenated along the channel dimension and fed into the DiT.
Our M2V design leverages this mask-guided conditional diffusion architecture, extending the conditioning to keyframe-based memory contexts for multi-shot generation.

\subsection{Problem Formulation}
\label{subsec:formulation}
Given a story script consisting of a sequence of textual descriptions for each shot $\mathcal{T} = \{t_i\}_{i=1}^{N}$, our goal is to generate a coherent multi-shot narrative video $\mathcal{V} = \{v_i\}_{i=1}^{N}$.
This task can be expressed as learning the conditional distribution:
\begin{equation}
    p_\Theta(\mathcal{V}\mid\mathcal{T}) = p_\Theta(v_{1:N}\mid t_{1:N}),
\end{equation}
which captures both the semantic relations among textual prompts and the visual coherence among generated shots.
To leverage the strong capability of existing single-shot video diffusion models, we decompose the joint distribution in an autoregressive form, which aligns with the natural progression of stories:
\begin{equation}
p_\Theta(\mathcal{V}\mid\mathcal{T})
= \prod_{i=1}^{N} p_\Theta(v_i \mid v_{1:i-1}, \mathcal{T}).
\end{equation}
However, since video frames contain massive temporal redundancy, conditioning directly on all past video shots $v_{1:i-1}$ is highly inefficient.
A simplified alternative is to first generate global consistent keyframes $\mathcal{K}=\{k_i\}_{i=0}^{N}$ with a consistent image generator $p_\psi$ and then independently synthesize each shot with an image-to-video model, \ie,
\begin{equation}
        p_{\Theta,\psi}(\mathcal{V}, \mathcal{K} \mid \mathcal{T})\approx 
        p_\psi(\mathcal{K} \mid \mathcal{T})
        \prod_{i=1}^{N} p_\Theta(v_i \mid t_i, k_i).
\end{equation}
However, this formulation lacks temporal awareness and inherently suffers from limited video context and rigid transitions, failing to capture the evolving narrative across shots.

Inspired by human memory, which selectively retains key visual impressions rather than full experiences, we introduce an explicit keyframe-based memory mechanism, where a compact memory $m_i$ stores key visual contexts up to shot $i$.
Let $\mathcal{M} = \{m_i\}_{i=0}^{N}$ denote the memory sequence.
By default, $m_0$ is empty and the first shot is generated solely from its textual description $t_1$, but our framework also support an alternative initialization if desired.
The joint distribution $p_{\Theta,\phi}(\mathcal{V}, \mathcal{M}\mid\mathcal{T})$ can be written as:
\begin{equation}
    p_\phi(m_0)\,
    \prod_{i=1}^{N}
    p_\Theta\big(v_i \mid t_i, m_{i-1}\big)\,
    p_\phi\big(m_i \mid m_{i-1}, v_i, t_i\big),
\end{equation}
where $p_\Theta$ parameterizes the shot-level video generator and $p_\phi$ denotes the memory update mechanism.

In our framework, $m_i$ is realized as a small set of keyframes extracted from previous shots, providing explicit visual anchors for subsequent generation.
The memory update is implemented as a deterministic function:
\begin{equation}
    m_i = f_\phi(m_{i-1}, v_i),
\end{equation}
yielding the simplified factorization:
\begin{equation}
p_\Theta(\mathcal{V}\mid\mathcal{T})
\approx \prod_{i=1}^{N} p_\Theta(v_i \mid t_i, m_{i-1}).
\label{eq:formulation}
\end{equation}
This formulation enables \textit{memory-based multi-shot generation}, where each shot is conditioned on both its textual description and an evolving memory that summarizes characters, scenes, and stylistic information from previous shots, thereby ensuring cross-shot consistency and narrative coherence throughout the entire video.

\subsection{Memory-to-Video (M2V)}
\label{subsec:m2v}

Based on the above formulation, we propose the Memory-to-Video (M2V) mechanism, which realizes the conditional distribution $p_\Theta(v_i \mid t_i, m_{i-1})$ via a memory-conditioned video diffusion model with negative RoPE shift.

\noindent\textbf{Memory as Condition.}
To achieve memory-conditioned video generation, we leverage the architecture design of pretrained image-conditioned video model Wan-I2V.
We first encode the memory frames into memory latents $z_m\in\mathbb{R}^{c\times f_m\times h\times w}$ using the same 3D VAE encoder $\mathcal{E}$ used by the base model, where each latent corresponds to one memory frame without temporal compression.
These memory latents are concatenated with latents encoded from zero-filled frames along the temporal dimension to form the conditional latent $z_c$.
A binary mask $M\in\{0,1\}^{s\times(f_m+f)\times h\times w}$ marks the preserved memory frames (1) and generative regions (0), guiding the DiT to generate only the unmasked frames.
Following the I2V model design, the noisy latent, the memory conditional latent, and the mask is concatenated together along channel dimension and fed into DiT for velocity prediction.

\noindent\textbf{Negative RoPE Shift.}
Incorporating memory frames into the video sequence raises a positional encoding issue: memory frames are not temporally continuous with the current shot but represent discrete keyframes summarizing past content.
To handle this, we leverage the generalizability of 3D RoPE and extend it by assigning negative frame indices to memory latents.
Specifically, for a shot with $f$ latent frames and $f_m$ memory frames, the temporal positional indices are defined as $\{-f_m S, -(f_m-1)S, \dots, -S, 0, 1, \dots, f-1\}$, where $S$ is a fixed offset indicating the temporal gap between memory latents and video latents.
This negative RoPE shift naturally embeds memory frames as preceding events, preserving the original temporal encoding of the current video starting from zero, which allows the DiT to seamlessly attend across previous and current contexts within a unified temporal space.

\noindent\textbf{Data Curation and Training.}
Unlike joint modeling methods that rely on multi-shot long sequences, our model can be effectively trained on single-shot videos.
To curate memory–video coherent data, we collect training data from: 
(1) visually related short clips grouped from cinematic single-shot video datasets by shot-level similarity, and (2) high-quality single-subject multi-scene short videos that naturally exhibit consistent identity.
During training, one specific clip within each group is selected as target, with frames from other clips randomly sampled as the memory.
The model is then trained to reconstruct the target video conditioned on the provided memory frames, following the rectified flow loss used in I2V model (Eq. \ref{eq:rf}).
At inference time, the memory part is discarded and only the newly generated segment is decoded as the video shot.

\subsection{Memory Extraction and Update}
\label{subsec:mfextract}

After training, the finetuned M2V model learns to retrieve relevant context from the memory bank following the prompt instruction, generating diverse shots with coherent characters and backgrounds. 
To apply our M2V model for multi-shot long video generation, we further design a memory update function $f_\phi$ that mimics human memory by preserving representative, meaningful moments and discarding redundancy. 
To this end, we propose a memory frame extraction strategy that selects semantically important and aesthetically reliable keyframes as memory.

\begin{figure*}[t]
    \vspace{-0.3cm}
    \centering
    \includegraphics[width=\linewidth]{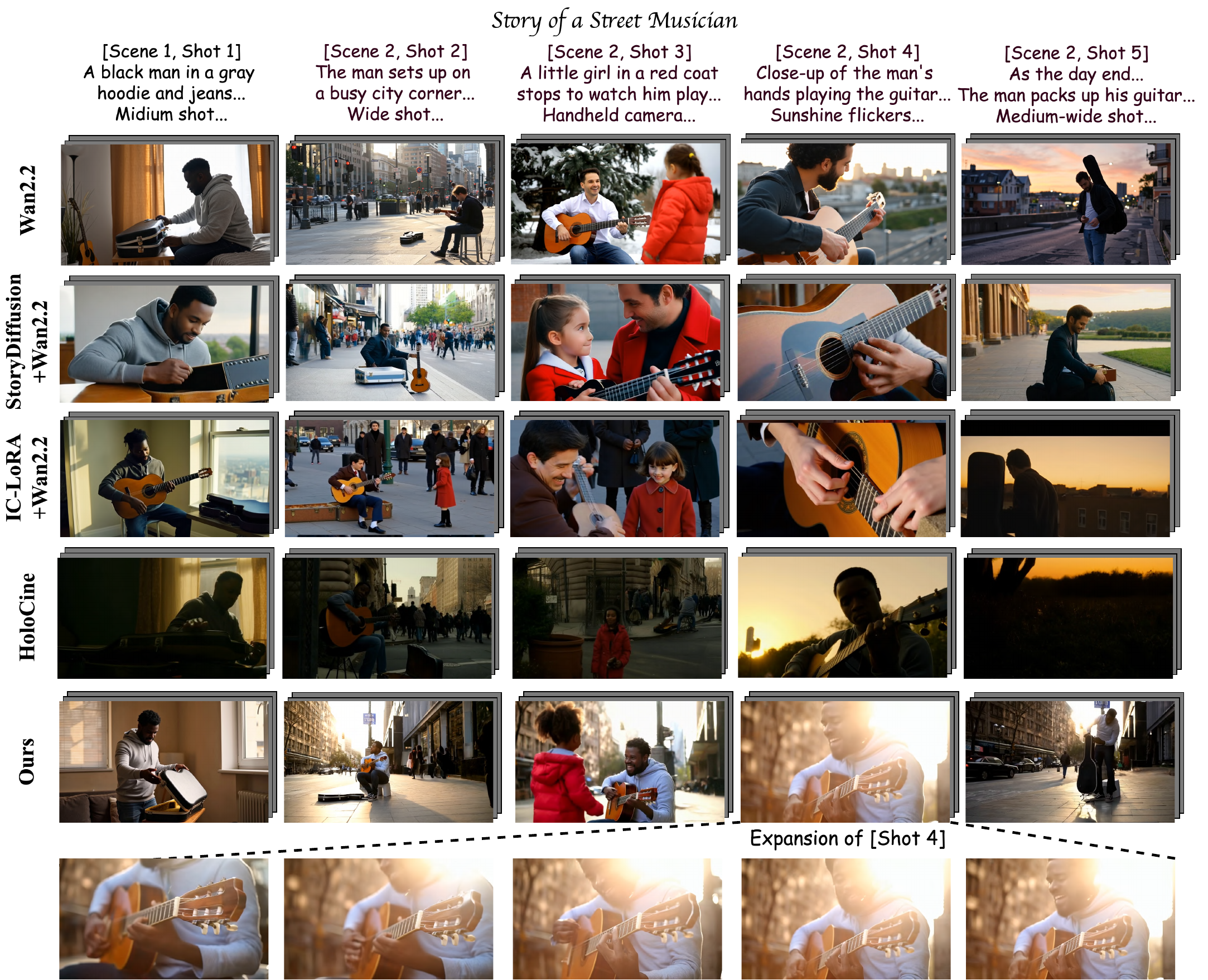}
    \caption{\textbf{Qualitative comparison.} Our~\ours~generates coherent multi-scene, multi-shot story videos aligned with per-shot descriptions. In contrast, the pretrained model and keyframe-based baselines fail to preserve long-term character and scene consistency, while HoloCine~\cite{meng2025holocine} exhibits noticeable degradation in visual quality.}
    \vspace{-0.3cm}
    \label{fig:cmp}
\end{figure*}

\noindent\textbf{Semantic Keyframe Selection.}
For each generated shot, we aim to identify a small set of distinct keyframes as memory.
To capture frame-wise semantics, we compute CLIP~\cite{clip} embeddings for all frames and sequentially select keyframes: the first frame is fixed, and each subsequent frame is compared with the latest keyframe via cosine similarity. 
A new keyframe is added when the similarity drops below a dynamic threshold, which starts low and increases if the number of selected frames exceeds a preset upper bound. 
This adaptive strategy removes redundancy while preserving diverse visual context for memory construction.

\noindent\textbf{Aesthetic Preference Filtering.}
While semantic selection effectively identifies distinct content, it does not guarantee image quality:
\eg, blurred or noisy frames from videos with large motion may be wrongly selected as keyframes, providing limited contextual guidance.
We address this by further filtering using the HPSv3~\cite{hpsv3} as aesthetic reward model,  ensuring clear and visually reliable memories.

\noindent\textbf{Multi-shot Long Video Generation.}
Integrating the above components, we use the M2V model for multi-shot generation with a dynamic memory bank that updates keyframes across shots. 
After generating each shot $v_i$, we compare its extracted keyframes with existing memories in CLIP~\cite{clip} space and add only semantically distinct ones, updating $m_{i-1}$ to $m_i$.
To prevent uncontrolled growth of memory bank, we adopt a hybrid \textit{memory-sink + sliding-window} strategy: early keyframes are fixed as long-term anchors preserving the global consistency, while recent ones form a short-term window capturing local dependencies.
If the capacity limit is reached, the oldest short-term memories are discarded.
If desired, human creators or large vision–language models can further review and refine the selected keyframes for finer story-specific control, although this option is not used in our results.

\subsection{Extension to MI2V and MR2V}
\label{subsec:m2vext}
As mentioned in Sec.~\ref{subsec:formulation}, our framework provides a flexible design space and can be seamlessly adaptable with other orthogonal video generation paradigms.

One important design choice is to combine M2V with I2V.
While the memory-conditioned paradigm effectively addresses cross-shot consistency, the transitions between shots may still appear unnatural when concatenated into a long video, and sometimes leading to non-causal motions.
A practical solution is to include a scene-cut indicator in the story script, allowing the script creator (LLM or human) to decide whether a scene cut should occur between each two adjacent shots.
If no cut is specified, the model reuses the last frame of the previous shot as the first frame of the next one, achieving smoother and more natural continuity.

Another application is to personalize the initialization of the memory state $m_0$.
For instance, users can provide character or background reference images as the initial memory, enabling customized multi-shot video generation.
\begin{table*}[t]
\centering
\caption{\textbf{Quantitative comparison on \textit{ST-Bench}.} The best and runner-up are in \textbf{bold} and \underline{underlined}. 
Note that the pretrained base model \textit{Wan2.2-T2V} independently generates each shot without consistency constraints, offering an upper bound reference for single-shot metrics.}
\small
\setlength{\tabcolsep}{6pt}
\resizebox{0.85\textwidth}{!}{%
\begin{tabular}{lcccccc}
\toprule
\textbf{Method} & 
\textbf{Aesthetic Quality↑} &
\multicolumn{2}{c}{\textbf{Prompt Following↑}} & 
\multicolumn{2}{c}{\textbf{Cross-shot Consistency↑}} \\
\cmidrule(lr){3-4} \cmidrule(lr){5-6}
& & \textbf{Global} & \textbf{Single-shot} & \textbf{Overall} & \textbf{Top-10 Pairs} \\
\midrule
Wan2.2-T2V & 0.6452 & 0.2174 & 0.2452 & 0.3937 & 0.4248 \\
\midrule
StoryDiffusion + Wan2.2-I2V & \underline{0.6085} & \underline{0.2288} & \textbf{0.2349} & 0.4600 & 0.4732 \\
IC-LoRA + Wan2.2-I2V & 0.5704 & 0.2131 & 0.2181 & 0.4110 & 0.4296 \\
HoloCine & 0.5653 & 0.2199 & 0.2125 & \underline{0.4628} & \underline{0.5005} \\
Ours & \textbf{0.6133} & \textbf{0.2289} & \underline{0.2313} & \textbf{0.5065} & \textbf{0.5337} \\
\bottomrule
\end{tabular}%
}
\label{tab:quantitative}
\end{table*}


\begin{figure*}[t]
    \centering
    \includegraphics[width=\linewidth]{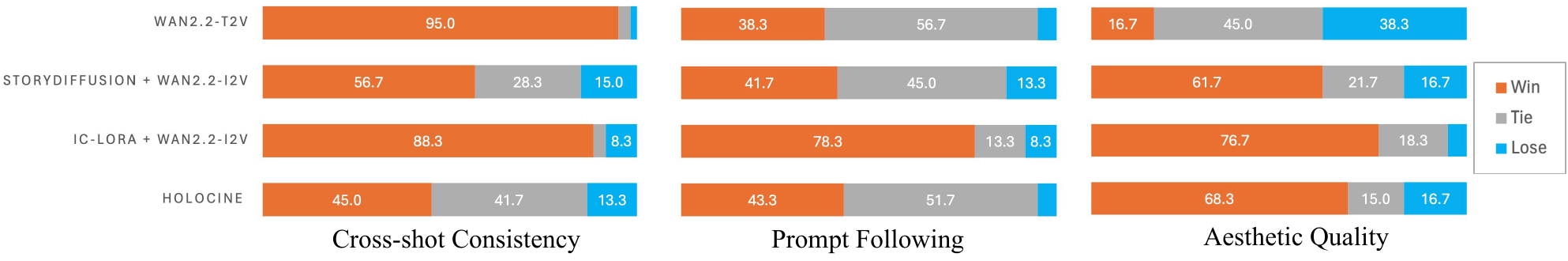}
    \caption{\textbf{User study.} Our method is consistently preferred over all baselines in most dimensions, highlighting its superior multi-shot consistency and narrative coherence. \textit{Win} indicates that users prefer our method over the baseline, \textit{Tie} indicates no significant preference, and \textit{Lose} indicates that users prefer the baseline.}
    \vspace{-0.3cm}
\label{fig:user}
\end{figure*}

\section{Experiments}
\label{sec:exp}

\subsection{Implementation Details}
Our framework is built upon the state-of-the-art open-source video generation model \textit{Wan2.2-I2V-A14B}~\cite{wan2025} with 14B active parameters.
We finetune it using a rank-128 LoRA applied to all linear layers in the DiT blocks, adding $\sim$0.7B active parameters..
The M2V model is trained on a curated dataset of 400K five-second single-shot videos, where each clip is matched with 1–5 semantically coherent videos.
During training, the memory length is randomly sampled from 1–10 frames, and the negative RoPE shift offset $S$ is set to $5$.
During inference, we use a memory sink size of $3$ frames, a per-shot keyframe limit of $3$, an initial semantic similarity threshold of $0.9$, and an aesthetic score threshold of $3.0$.

\subsection{ST-Bench}
Multi-shot long video storytelling remains a relatively new and underexplored research area, with no standard evaluation benchmark.
The most relevant open benchmark, \textit{ViStoryBench}~\cite{zhuang2025vistorybench}, focuses on storyboard image generation, making it unsuitable for our video generation task.
To comprehensively evaluate our method, we establish a new multi-scene, multi-shot story video generation benchmark, termed \textit{ST-Bench}.
We prompt GPT-5~\cite{gpt5} to create 30 long story scripts spanning diverse styles, each containing a story overview, 8–12 shot-level text prompts, corresponding first-frame prompts (only for two-stage keyframe-based methods), and scene-cut indicators.
In total, \textit{ST-Bench} provides 300 detailed video prompts describing characters, scenes, dynamic events, shot types, and possibly camera movements.
More details about \textit{ST-Bench} are presented in Appendix~\ref{app:bench}.
We hope this benchmark can facilitate future research in long-form story video generation.

\begin{figure}[t]
    \vspace{-0.5cm}
    \centering
    \includegraphics[width=0.8\linewidth]{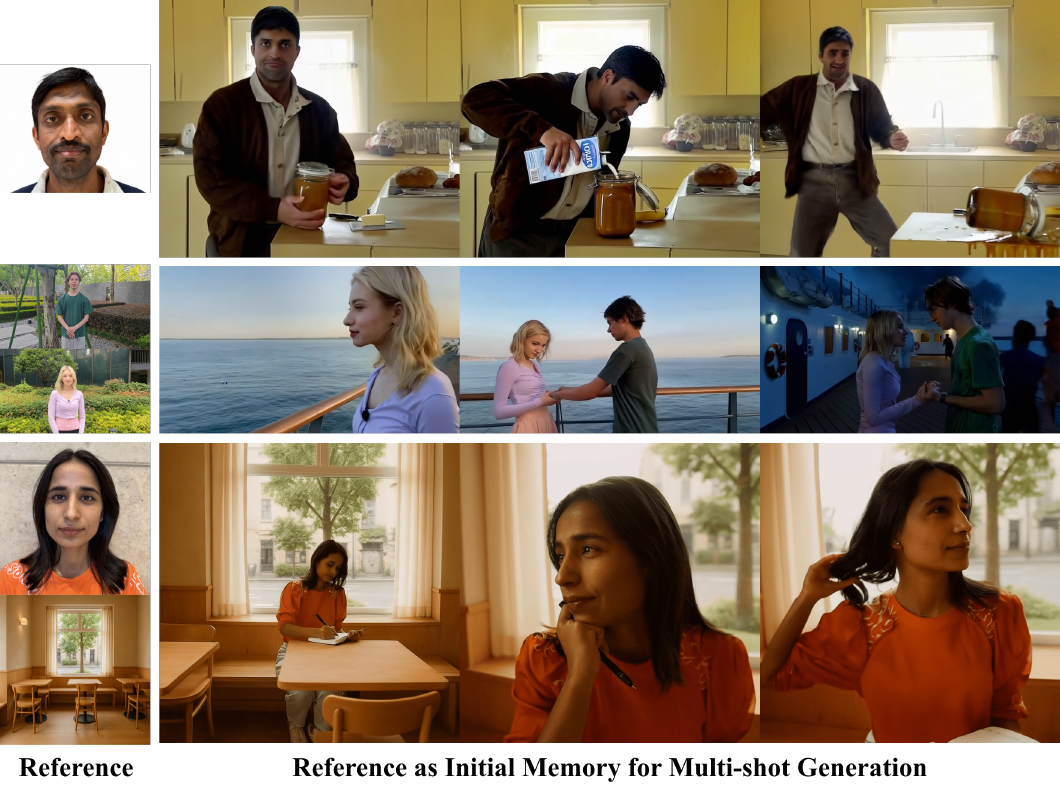}
    \vspace{-0.2cm}
    \caption{\textbf{MR2V results.} \ours~enables customized story video generation by using reference images as the initial memory. The real-person reference images were used with proper consent from the individuals involved.}
    \vspace{-0.3cm}
    \label{fig:mr2v}
\end{figure}

\subsection{Evaluation}
\noindent\textbf{Baseline Setup.} 
To demonstrate the superiority of~\ours, we follow the evaluation protocol of HoloCine~\cite{meng2025holocine} and compare \ours\ with three representative paradigms of multi-shot long video generation: (1) a pretrained video diffusion model, \textit{Wan2.2-T2V-A14B}, applied independently to each shot, serving as a reference for single-shot quality; (2) two-stage keyframe-based approaches, StoryDiffusion~\cite{Zhou2024StoryDiffusion} and IC-LoRA~\cite{Huang2024ICLora}, with keyframes expanded using \textit{Wan2.2-I2V-A14B}; and (3) the state-of-the-art joint multi-shot model HoloCine~\cite{meng2025holocine}, which finetunes \textit{Wan2.2-T2V-A14B} for holistic one-minute video generation. For compatibility, prompts in \textit{ST-Bench} are converted to HoloCine’s required format using GPT-5.


\noindent\textbf{Qualitative Results.}
We provide a qualitative comparison to illustrate the superiority of our method. 
As shown in Fig.~\ref{fig:cmp}, our~\ours~generates coherent multi-shot story videos that closely follow the per-shot descriptions while maintaining high visual fidelity. 
The character’s identity, appearance and outfits remain consistent across different scenes and shots.
Notably, in [Shot~5], our model effectively retrieves contextual information from [Shot~2], producing highly consistent street scenes even after multiple shot transitions.
In contrast, the pretrained model and keyframe-based baselines fail to preserve long-term consistency (e.g., mismatched character identity, altered outfits, inconsistent street appearance), and HoloCine exhibits notable degradation in visual quality.
Additional qualitative examples are provided in Appendix~\ref{app:results}.

\noindent\textbf{Quantitative Results.}
Following previous works~\cite{Guo2025LCT, meng2025holocine}, we evaluate all methods across three aspects: (1) Aesthetic Quality is measured using the LAION aesthetic predictor adopted in VBench~\cite{huang2023vbench}, reflecting visual appeal including color harmony, realism, and naturalness. (2) Prompt Following is assessed with ViCLIP~\cite{wang2023internvid} by comparing generated videos with the story scripts; the global score computes the cosine similarity between the entire multi-shot video and the story overview, while the single-shot score compares each shot with its corresponding prompt. (3) Cross-shot Consistency is computed as the mean ViCLIP similarity across all shot pairs. Since different shots may depict distinct characters or scenes, we additionally report a Top-10 Consistency score by averaging the ten most relevant shot pairs selected via their prompt feature similarities.

Tab.~\ref{tab:quantitative} reports the quantitative results of all methods on \textit{ST-Bench}.
Our method surpasses all baselines in Cross-shot Consistency metrics by a large margin, outperforming the base model by 28.7\% and the previous state-of-the-art HoloCine by 9.4\% in overall consistency.
This highlights the strength of our memory-conditioned design in maintaining long-range coherence across shots.
Our approach also attains strong performance in both Aesthetic Quality and Prompt Following, achieving the highest aesthetic score and the best global semantic alignment among all consistent video generation methods.
Although the single-shot prompt following score is slightly lower, this is expected under our MI2V setting, which introduces additional control to achieve smoother and more natural shot transitions, and may slightly constrain single-shot alignment.

\noindent\textbf{User Study.}
Video storytelling is a complex and human-centered task.
Existing metrics cannot fully capture aspects such as character-level cross-shot consistency or the naturalness of shot transitions. 
Therefore, we additionally conduct a comprehensive user study, where human evaluators compare our generated results against each baseline in pairwise settings.
As shown in Fig.~\ref{fig:user}, our results are strongly preferred over all baselines across most evaluation dimensions. 
Although the pretrained Wan2.2 model performs well on single-shot quality due to its independent shot generation setting, it lacks consistency mechanisms and performs significantly worse on other dimensions.

\noindent\textbf{Extended Application.}
Thanks to the flexible design of our memory-conditioned framework, \ours~can also support customized story generation by treating reference images as the initial memory, as shown in Fig.~\ref{fig:mr2v}. 
Compared with conventional reference-based methods, our approach enables natural narrative progression by both preserving newly generated scenes and selectively leveraging relevant context stored in memory. 
Our framework is also compatible with other specialized reference-preserving techniques, which is not the focus of our work.

\begin{figure}[t]
    \centering
    \includegraphics[width=0.95\linewidth]{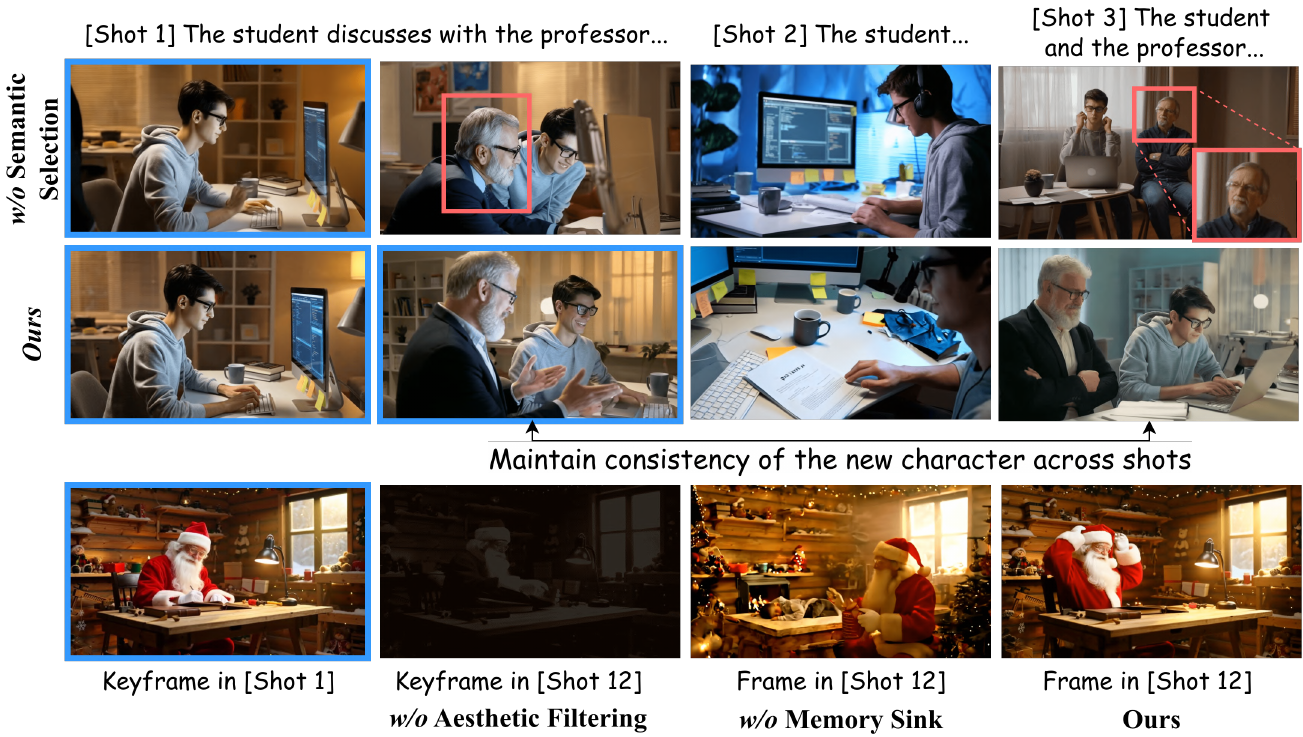}
    \caption{\textbf{Ablation study.} Top: our method effectively preserves newly emerged character consistency. Bottom: our method can maintain long-term visual fidelity. The selected keyframe in [Shot~1] is highlighted in blue box.}
    \vspace{-0.3cm}
\label{fig:ablation}
\end{figure}

\begin{table}[t]
\centering
\vspace{-0.2cm}
\caption{\textbf{Ablation study.} We study the effectiveness of each proposed techniques in~\ours.}
\small
\setlength{\tabcolsep}{6pt}
\resizebox{0.85\linewidth}{!}{%
\begin{tabular}{lcccccc}
\toprule
\textbf{Method} & 
\textbf{Aesthetic Quality↑} &
\multicolumn{2}{c}{\textbf{Prompt Following↑}} & 
\multicolumn{2}{c}{\textbf{Cross-shot Consistency↑}} \\
\cmidrule(lr){3-4} \cmidrule(lr){5-6}
& & \textbf{Global} & \textbf{Single-shot} & \textbf{Overall} & \textbf{Top-10 Pairs} \\
\midrule
\textit{w/o} Semantic Selection & 0.6076 & 0.2257 & 0.2295 & 0.4878 & 0.5287 \\
\textit{w/o} Aesthetic Filtering & 0.6018 & 0.2251 & 0.2313 & 0.4844 & 0.5330 \\
\textit{w/o} Memory Sink & 0.6093 & 0.2277  & \textbf{0.2330}  & 0.4891  &  0.5241 \\
Ours & \textbf{0.6133} & \textbf{0.2289} & 0.2313 & \textbf{0.5065} & \textbf{0.5337} \\
\bottomrule
\end{tabular}%
}
\vspace{-0.2cm}
\label{tab:ablation}
\end{table}

\subsection{Ablation Study}
We conduct ablation studies to validate the effectiveness of our proposed techniques.
To evaluate our semantic keyframe selection, we compare it with a naive strategy that always selects the first generated frame of each shot as memory. 
As shown in the top example of Fig.~\ref{fig:ablation}, this naive choice fails to capture newly introduced characters (e.g., the professor) within video shots, leading to inconsistent appearance and outfits.
To assess the importance of aesthetic filtering, we remove this step during memory extraction. 
The bottom example in Fig.~\ref{fig:ablation} shows that without aesthetic filtering, the semantic selection becomes noise-sensitive and may include low-quality or uninformative frames, resulting in the lowest aesthetic score in Tab.~\ref{tab:ablation}.
Finally, to examine the memory sink mechanism, we replace it with a full sliding-window strategy once the memory bank is full. 
This leads to degradation in long-range visual fidelity, whereas our method preserves high quality and consistency in minute-scale generation.
Overall, as shown in Tab.~\ref{tab:ablation}, our method achieves the best performance across most metrics, with only a slight drop in single-shot prompt following due to the additional initial-memory constraints.

\section{Conclusion}
\label{sec:conclus}

\begin{figure}[t]
    \vspace{-0.2cm}
    \centering
    \includegraphics[width=0.8\linewidth]{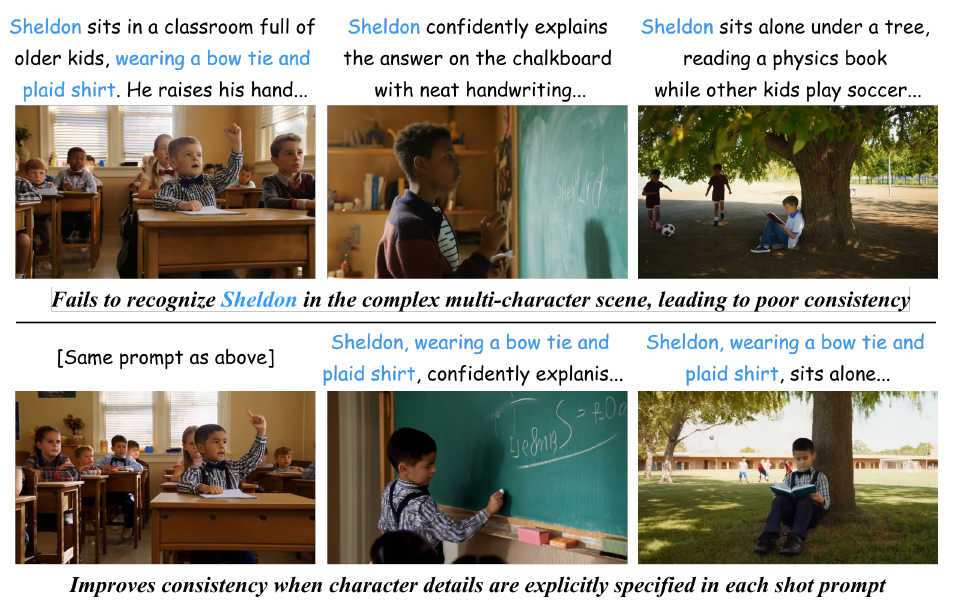}
    \vspace{-0.3cm}
    \caption{Limitation. Top: Our method may struggle to preserve consistency in complex multi-character scenarios where pure visual memory becomes ambiguous. Bottom: Explicitly providing character details in each shot prompt will mitigate the problem.}
    \vspace{-0.3cm}
\label{fig:limitation}
\end{figure}

We presented \ours, a paradigm for coherent multi-shot long video storytelling. 
To this end, we introduced a Memory-to-Video (M2V) framework, which augments single-shot diffusion models with explicit visual memory through latent concatenation, negative RoPE shifts, and lightweight LoRA fine-tuning. 
Combined with a delicate memory extraction and update strategy, \ours~enables shot-by-shot synthesis of minute-long narrative videos. 
Extensive experiments show that~\ours~outperforms state-of-the-art methods, delivering robust cross-shot coherence while preserving aesthetic quality and prompt adherence. 
While effective, \ours~may still struggle in complex multi-character scenarios where pure visual memory becomes ambiguous, and in achieving fully smooth transitions when adjacent shots exhibit large motion discrepancies. 
Future work will explore more structured, entity-aware memory representations and improved transition modeling to better support rich and fluid narratives.

\clearpage

\bibliographystyle{plainnat}
\bibliography{main}

\clearpage

\beginappendix

\section{More Qualitative Results}
\label{app:results}



\begin{figure*}[t]
    \centering
    \includegraphics[width=\linewidth]{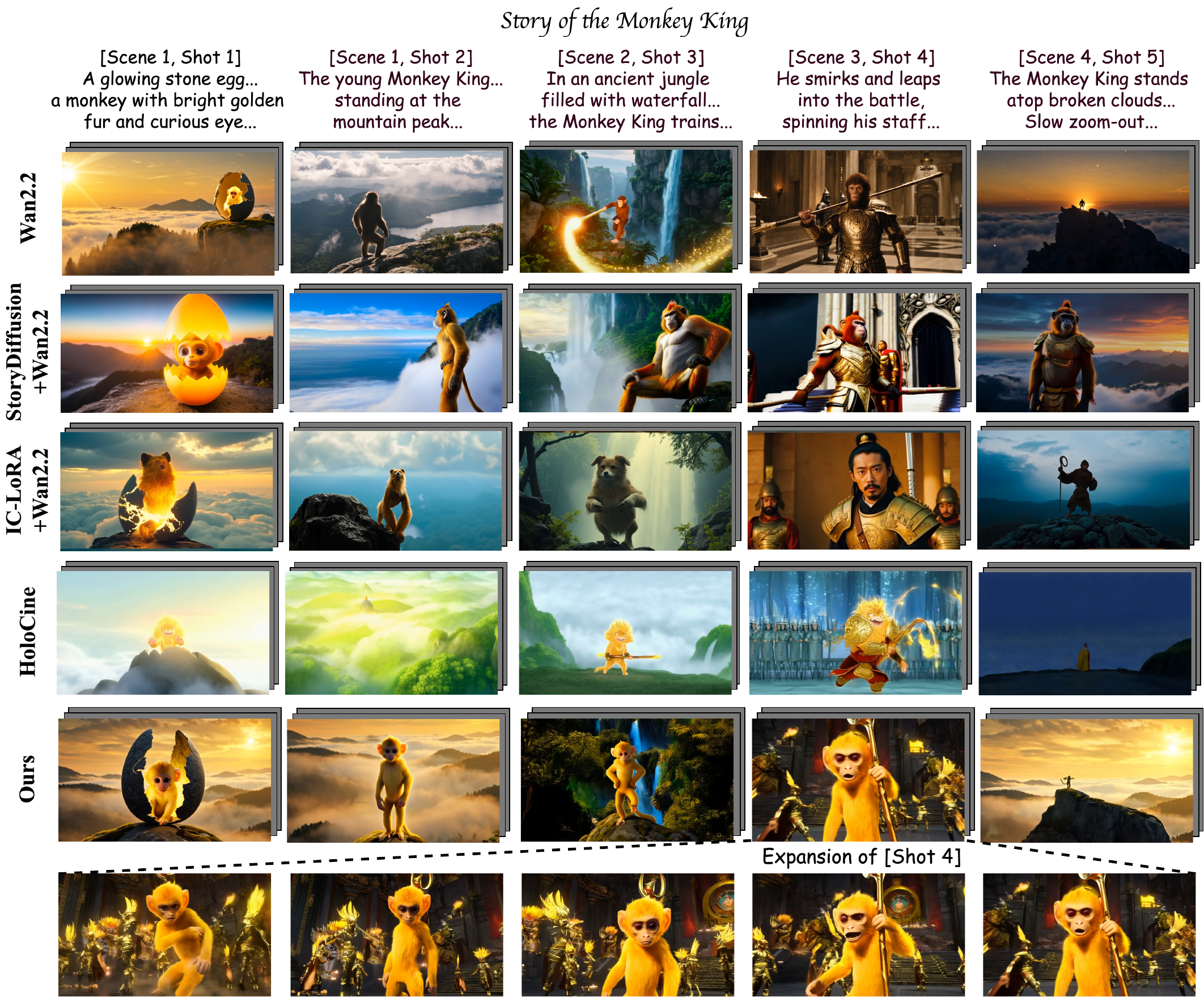}
    \caption{Additional qualitative comparison.}
    \label{fig:cmp2}
\end{figure*}

\begin{figure*}[t]
    \centering
    \includegraphics[width=\linewidth]{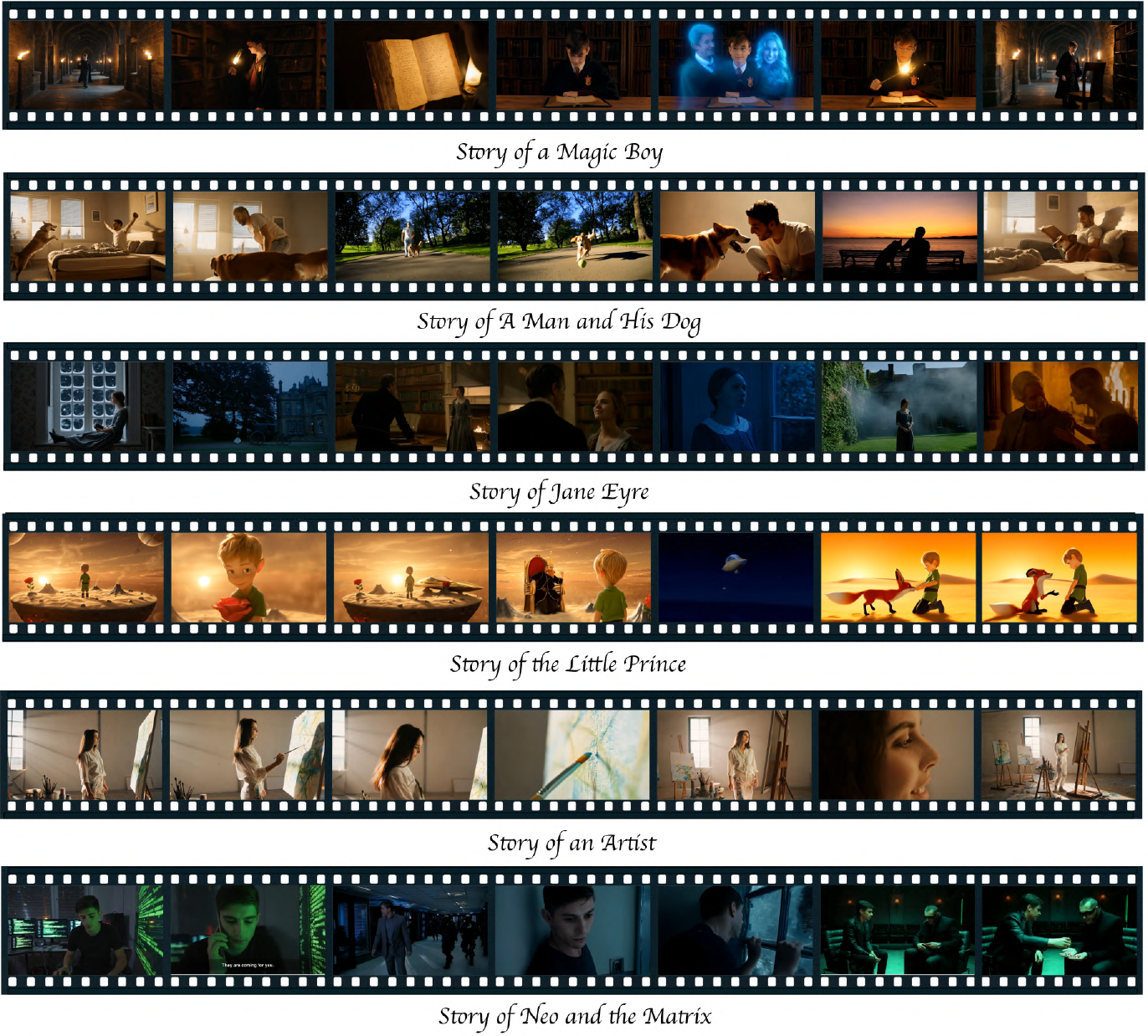}
    \caption{More qualitative results.}
    \label{fig:qualitative}
\end{figure*}

Here we present additional qualitative results to further demonstrate the effectiveness of~\ours.
Fig.~\ref{fig:cmp2} provides another example illustrating the superiority of our method across diverse visual styles.
Fig.~\ref{fig:qualitative} shows more results generated by our approach, further highlighting its versatility in story video generation.

\section{More Details of ST-Bench}
\label{app:bench}

Here we provide additional details on the construction of \textit{ST-Bench}. 
As described in the main paper, \textit{ST-Bench} consists of 30 long-form story scripts generated using GPT-5, yielding 300 video prompts in total. 
Each story adheres to a unified JSON schema that includes a short story overview, 1-4 scenes, 8-12 shot-level video prompts, corresponding first-frame prompts (only for keyframe-based two-stage methods), and cut indicators specifying whether a shot begins with a hard scene transition or continues smoothly from the previous one. 
To ensure high-quality and model-friendly prompts, we design a structured system instruction that restricts each shot description to 1-4 concise sentences, emphasizing character appearance, simple motions, scene layout, mood, and lightweight camera guidance, as shown in Fig.~\ref{fig:sysprompt}.
The resulting benchmark spans diverse styles, ranging from realistic to fairy-tale stories, from ancient to modern settings, and from Western to Eastern cultural aesthetics, ensuring that different generation paradigms receive standardized and well-structured inputs.
A complete example story script (excluding first-frame prompts) is provided in Fig.~\ref{fig:script} to illustrate the format used in \textit{ST-Bench}. 
All story scripts will be released to support future research in this area.

\section{Limitation and Future Work}
One major limitation of our visual memory mechanism is the ambiguous retrieval problem.
Limited by the architecture design of our base model \textit{Wan2.2}, which uses cross-attention-based DiT instead of more flexible MMDiT~\cite{sd3, flux2024, Kong2024HunyuanVideoAS}, our memory is purely visual and does not incorporate textual meta information.
In other words, the memory update function $m_i = f_\phi(m_{i-1}, v_i)$ (Eq.~8) does not include the textual information $t_i$, as required in the standard formulation in Eq.~7.
As a result, in complex multi-character scenarios, the model may fail to retrieve the correct context from memory given the current shot prompt, leading to inconsistent character appearance across shots, as illustrated in the top row of Fig.~\ref{fig:limitation}.
A simple mitigation is to provide more explicit character descriptions in each shot prompt, helping the model match the intended memory, as shown in the bottom row of Fig.~\ref{fig:limitation}. 
In the future, we plan to explore more structured, entity-aware memory representations to address this limitation more fundamentally.

Another minor limitation lies in achieving fully smooth shot transitions.
Although our autoregressive shot generation process and MI2V design significantly alleviate the rigid transitions in prior keyframe-based approaches, the single-frame connection mechanism does not convey video speed information. 
As a result, when two adjacent MI2V-connected shots differ significantly in motion speed, the transitions may still appear unnatural, as seen in some provided video results.
In future work, we plan to achieve smoother transitions by overlapping more frames across consecutive shots where no scene-cut is intended.

\begin{figure*}[t]
\centering
\begin{tcolorbox}[
    colback=black!3,
    colframe=black!20,
    boxrule=0.4pt,
    arc=2pt,
    left=6pt,right=6pt,top=6pt,bottom=6pt,
    width=\textwidth
]
\footnotesize
\ttfamily
You are an expert director of story videos. Your task is to design a story script about [\textellipsis]. 
The script should be divided into 1-4 scenes, each maintaining consistent background and character design, and containing no more than 12 prompts in total.

\medskip

Each prompt corresponds to a five-second video clip, so avoid overly complex text rendering, extreme motions, or audio-dependent effects. The overall story should remain simple, clear, and easy to follow.

\medskip

Your output must follow the JSON format shown in the example:

\texttt{[ ... an example json story script here ... ]}

\medskip

\textbf{Field Instructions:}

\begin{itemize}
    \item \textbf{story\_overview}: A concise summary of the whole story.
    \item \textbf{scene\_num}: Sequential index of the scene.
    \item \textbf{cut}: Whether this prompt starts with a scene cut:
    \begin{itemize}
        \item \texttt{"True"}: a new cut.
        \item \texttt{"False"}: continue smoothly from the last frame of the previous prompt. Must ensure the two adjacent prompts can be naturally concatenated into a smooth continuous clip.
        \item The first prompt in the story must always have \texttt{"True"}.
    \end{itemize}
    \item \textbf{video\_prompts}: A list of text-to-video prompts forming the story beats within the scene. Prompts should reflect natural, smooth, and logical story progression.
\end{itemize}

\medskip

\textbf{What Each Video Prompt Should Describe (if relevant):}

\begin{itemize}
    \item \textbf{Characters}: appearance, attire, age, style.
    \item \textbf{Actions \& interactions}: Motion, gestures, expressions, eye contact, simple physical actions.
    \item \textbf{Scene \& background}: Indoor/outdoor location, props, layout, lighting, environment details.
    \item \textbf{Atmosphere \& mood}: Emotional tone, colors, aesthetic feeling.
    \item \textbf{Camera \& editing}: Shot type (\eg, close-up/medium/wide), simple camera movement, transitions.
\end{itemize}
The prompts should be concise but sufficiently detailed (1-4 sentences). 

\medskip

Return only a valid JSON story script.
\end{tcolorbox}
\caption{System prompt used for generating structured multi-shot story scripts in \textit{ST-Bench}.}
\label{fig:sysprompt}
\end{figure*}

\begin{figure*}[t]
\centering
\begin{tcolorbox}[
  colback=black!3,
  colframe=black!15,
  boxrule=0.3pt,
  arc=2pt,
  left=6pt,right=6pt,
  top=3pt,bottom=3pt,
  width=\textwidth
]
\footnotesize
\begin{lstlisting}[language=json,breaklines=true,frame=none]
{
  "story_name": "The Street Musician",
  "story_overview": "A humble black street musician finds beauty and purpose in everyday life through his music. From performing on a busy city street to sharing a smile with a child and returning home to play quietly under the soft glow of evening light, the story captures moments of sincerity, resilience, and the quiet joy of self-expression.",
  "scenes": [
    {
      "scene_num": 1,
      "video_prompts": [
        "Early morning in a small apartment. A black man in his 30s, wearing a simple gray hoodie and jeans, carefully opens an old guitar case on his small table. The room is lit by soft morning light from the window, filled with warm, earthy tones. Medium shot focusing on his calm expression as he tunes the strings.",
        "He stands by the window, strumming softly, looking out at the awakening city streets below. The sound of the strings fills the quiet air. Gentle camera pan from the window view to his face, capturing a sense of peace and purpose before he heads out."
      ],
      "cut": [true, false]
    },
    {
      "scene_num": 2,
      "video_prompts": [
        "The man sets up on a busy city corner, placing his guitar case open on the ground. People pass by - some glance, others hurry. He sits on a small stool and begins playing. Wide shot with bustling urban background and sunlight reflecting off buildings.",
        "A little girl in a red coat stops to watch him play, her eyes wide with curiosity. The man smiles warmly, slightly nodding to her as he continues strumming. Smooth transition from previous shot; handheld camera capturing the authenticity of the moment.",
        "Close-up of the man's hands moving gracefully on the guitar strings, the rhythm blending with the ambient city sounds. The sunlight flickers across the instrument, symbolizing hope. Shallow focus emphasizing texture and emotion."
      ],
      "cut": [true, false, true]
    },
    {
      "scene_num": 3,
      "video_prompts": [
        "As the day ends, the street grows quieter. The man packs up his guitar, pausing to look at the coins in his case - not much, but enough to bring a faint smile. The sky turns orange and pink above the buildings. Medium-wide shot with gentle dolly back.",
        "He walks home through narrow alleys lit by warm streetlights, guitar slung over his shoulder. His steps are slow but peaceful, matching the fading hum of the city. Smooth continuation from last frame with soft ambient background.",
        "At night, back in his apartment, he sits by the window again, softly playing the guitar under a dim lamp. The camera slowly zooms out through the window, showing the city glowing beyond. The mood is calm and introspective, ending on a gentle fade to black."
      ],
      "cut": [true, false, true]
    }
  ]
}
\end{lstlisting}
\end{tcolorbox}
\caption{Example of story script in \textit{ST-Bench}.}
\label{fig:script}
\end{figure*}

\end{document}